\documentclass[a4paper,fleqn]{cas-sc}

\usepackage{graphicx}
\usepackage{multirow}
\usepackage{amsmath,amssymb,amsfonts}
\usepackage{amsthm}
\usepackage{mathrsfs}
\usepackage[title]{appendix}
\usepackage{xcolor}
\usepackage{textcomp}
\usepackage{manyfoot}
\usepackage{booktabs}
\usepackage{algorithm}
\usepackage{algorithmicx}
\usepackage{algpseudocode}
\usepackage{listings}
\usepackage{subcaption}
\usepackage{tikz}
\usepackage{pgfplots}
\usepackage[authoryear,longnamesfirst]{natbib}
\usetikzlibrary{shapes, arrows}
\usetikzlibrary{calc}

\def\tsc#1{\csdef{#1}{\textsc{\lowercase{#1}}\xspace}}
\tsc{WGM}
\tsc{QE}
\tsc{EP}
\tsc{PMS}
\tsc{BEC}
\tsc{DE}

\begin{document}
\let\WriteBookmarks\relax
\def\floatpagepagefraction{1}
\def\textpagefraction{.001}
\shorttitle{Improving Trip Mode Choice Modeling Using Ensemble Synthesizer (ENSY)}
\shortauthors{Parsi et~al.}

\title [mode = title]{Improving Trip Mode Choice Modeling Using Ensemble Synthesizer (ENSY)}

\author[1]{Amirhossein Parsi}
\ead{amir.parsi78@sharif.edu}

\affiliation[1]{organization={Department of Civil Engineering, Sharif University of Technology},
                addressline={Azadi}, 
                city={Tehran},
                postcode={1458889694}, 
                country={Iran}}

\author[1]{Melina Jafari}
\ead{melina.jafari@sharif.edu}

\author[1]{Sina Sabzekar}
\ead{sina.sabzekar@sharif.edu}

\author[1]{Zahra Amini}
\cormark[1]
\ead{zahra.amini@sharif.edu}

\cortext[cor1]{Corresponding author}

\begin{abstract}
Accurate classification of mode choice datasets is crucial for transportation planning and decision-making processes. However, conventional classification models often struggle to adequately capture the nuanced patterns of minority classes within these datasets, leading to suboptimal accuracy. In response to this challenge, we present Ensemble Synthesizer (ENSY) which leverages probability distribution for data augmentation, a novel data model tailored specifically for enhancing classification accuracy in mode choice datasets. In our study, ENSY demonstrates remarkable efficacy by nearly quadrupling the F1 score of minority classes and improving overall classification accuracy by nearly 3\%. To assess its performance comprehensively, we compare ENSY against various augmentation techniques including Random Oversampling, SMOTE-NC, and CTGAN. Through experimentation, ENSY consistently outperforms these methods across various scenarios, underscoring its robustness and effectiveness.
\end{abstract}

\begin{keywords}
Oversampling \sep Guassian Mixture Model \sep Cumulative Distribution Function \sep CTGAN \sep SMOTE-NC \sep Data Augmentation
\end{keywords}

\maketitle

\section{Introduction}

In travel behavior research, understanding, and predicting travel mode choices have a pivotal role in the travel demand forecasting process. Factors influencing travelers' mode choices range from tangible elements like distance and travel time to intangibles like safety, reliability, and aesthetics \citep{Birr2018}. As advancements in technology usher in new modes of transportation, such as ride-sourcing, autonomous vehicles, and electric scooters, and with the increasing availability of diverse travel options, data and information sources will be altered. This calls for an evolution in the field of mode choice prediction modeling. 

Traditionally, mode choice prediction has been dominated by discrete choice models (DCMs) rooted in the principles of random utility maximization. While these models provide interpretability, they necessitate extensive efforts in specification and estimation, often requiring segmentation of travel markets based on various explanatory variables \citep{Hensher2015}, \citep{De_Ortuzar2011}, \citep{Richards2019}, \citep{Hillel2021}. Enter machine learning (ML) algorithms, which present a paradigm shift by avoiding rigid assumptions about data structures. ML techniques have entered the realm of mode choice modeling, offering a more flexible and efficient approach to understanding the nuances of travel behavior \citep{Hillel2021}, \citep{Wang2021}, \citep{Wang2019}, \citep{Pulugurta2013}.

In transportation research, datasets often exhibit a skewed distribution of classes, with some modes being significantly more prevalent than others. Thus, the challenges of imbalanced datasets in the context of mode choice prediction present a critical obstacle for accurate model performance. As ML applications become more prevalent, the issue of class imbalance becomes more pronounced, with traditional methods leaning towards the majority class and neglecting minority classes \citep{Mohammed2020}. To address this, several methods have emerged, including data augmentation techniques such as Synthetic Minority Over-sampling Technique (SMOTE) \citep{Chawla2002} and Generative Adversarial Networks (GANs) \citep{Aziira2020}. Moreover, due to the nature of mode choice problem datasets, handling categorical features is inevitable; which is more challenging than merely numerical datasets.

Each of the data augmentation methods aims to balance the class distribution, enhancing the robustness and accuracy of machine learning models in predicting travel behavior and contributing to a more comprehensive understanding of the diverse aspects of mode choice in modern transportation contexts \citep{Rezaei2022}. However, there is no one-size-fits-all solution to overcome class imbalance obstacles. Researchers have put extensive efforts into finding a solution, with most efforts proving fruitless \citep{Rezaei2022}, \citep{Diallo2022}, \citep{Li2021}, \citep{Chen2023}. Therefore, a new method is proposed in this paper to overcome the failure in improving the accuracy of minor classes in mode choice prediction. The method called ENSY, utilizes the probability distribution of the existing dataset to generate synthetic data points. Subsequently, a classifier assesses the generated data points for quality, and finally, the original dataset is augmented by incorporating valid synthetic data points.

In the course of this paper, the London Passenger Mode Choice dataset \citep{Hillel2018} and the Korea Transport Database \citep{KTDB2016} have undergone examination, and the aforementioned strategy is employed. The results from ENSY are compared to the results from previous data augmentation methods. Extreme Gradient Boosting, Random Forest, and Neural Networks are employed and the performance of models on both raw and augmented datasets is systematically compared.

\section{Related Works}
\subsection{Research on Mode Choice Classification}
Travel mode choice modeling is one of the most studied areas in travel behavior research and it is a crucial step in the travel demand forecasting process. The process of travel demand forecasting consists of the four-step model, including trip generation, trip distribution, modal split, and trip assignment \citep{Kadiyali2013}. 

Although several methods have been utilized for mode choice modeling, the field has been dominated for many years by the application of DCMs, including the Multinomial Logit (MNL), Nested Logit (NL), and Mixed Logit (MXL) models \citep{Zhao2020}. \citep{McFadden1974} pioneered the use of the MNL model in travel behavior modeling. In recent years, ML methods have gained traction in mode choice modeling. Notable ML techniques applied include Decision Trees, Neural Networks, and Support Vector Machines, and by combining some ML concepts such as ensemble methods, the prediction power has been enhanced \citep{Hillel2021}, \citep{Wang2021}, (Wang 2019), \citep{Pulugurta2013}, \citep{Zhang2020}, \citep{Rasouli2014}. 

\citep{Wang2018} compared the performance of Extreme Gradient Boosting (XGB) with traditional MNL models. Interestingly, while both XGB and MNL models demonstrated high prediction accuracy for travel mode choices, challenges arise when predicting choices involving cycling, which constitutes a small share of the dataset. 
\citep{Sekhar2016} explored the efficacy of a Random Forest (RF) Decision Tree mode choice model, showcasing its superiority over MNL models. This RF model not only achieved higher prediction accuracy but also demonstrated efficiency on large databases, emphasizing benefits such as accurate classification, scalability to large datasets, and internal error estimation.

In a study by Richards and Zill \citep{Richards2019}, the effectiveness of machine learning techniques in addressing the mode choice problem was assessed, with Gradient Boosting emerging as the top performer among the models considered. However, across this research and others, the issue of minority classes posed challenges due to severe class imbalance.

Consequently, XGB appears to outperform numerous ML classifiers, with the impact of class imbalance evident in predicting smaller shares, presenting a challenge that has seen limited efforts for resolution.

\subsection{Approaches to Handle Class Imbalance in Mode Choice Prediction}
As mentioned before, the class imbalance problem in ML occurs when skewed class distributions hinder classifiers from effectively learning information in minority classes, resulting in suboptimal performance. Four mainstream methods are explored to balance class distribution \citep{Majeed2023}. 

\subsubsection{Data-Level Methods}
These methods involve undersampling the majority classes or oversampling the minority classes, with one approach being a combination of both to enhance learning and generalization \citep{Chaipanha2022}, \citep{Menardi2014}.
\citep{Chen2023} systematically investigated the fusion of statistical and ML methods, with six Over/Under-Sampling (OUS) techniques. The examination revealed that while prediction models using the original dataset demonstrated superior aggregate prediction performance, the majority of OUS techniques proved advantageous in enhancing the disaggregate prediction performance of machine learning models. Notably, Random Under-Sampling (RUS) and oversampling techniques exhibited significant improvements in predicting minority modes while preserving overall prediction performance and model interpretability.

\subsubsection{Algorithm-Level Methods}
These methods modify ML model workflows, employing techniques like guiding Support Vector Machine hyperplanes and designing objective functions \citep{Batuwita2010}, \citep{Edward2003}. \citep{Qian2021} presented a novel approach to address imbalanced mode choice data using an adjustable kernel-based Support Vector Machine (SVM) classification model. The proposed method employs the likelihood-ratio chi-square test and weighting measures for optimal kernel function selection, incorporating a transformation function to expand class limits and rectify irregular boundaries. Notably, the SVM with Adjustable Kernel model significantly enhances prediction accuracy, improving from 82.33\% to an impressive 99.81\% for the largest sample size. However, for the smallest category (the motorcycle/moped), the developed models failed to improve the accuracy.

\subsubsection{Cost-Sensitive Methods}
These methods minimize misclassification costs, assigning higher costs to the minority classes \citep{Wang2010}. \citep{Kim2021} assessed that RF and XGB exhibit superior performance compared to Artificial Neural Networks. Despite efforts to address challenges related to imbalanced datasets by applying class-specific weights during ML model training, all models displayed suboptimal performance in predicting the minority class, specifically the choice of cycling.

\subsubsection{Ensemble Methods}
These methods involve training multiple classifiers to enhance accuracy in imbalanced scenarios \citep{Song2023}. (Wang 2019) established an empirical benchmark for predicting travel mode choice by employing 86 machine learning classifiers across 14 model families. The analysis reveals that ensemble models, specifically Boosting, Bagging, and RF, outperform other classifiers. Notably, Bagging attains the highest prediction accuracy among the 14 model families. Although, using ensemble learning led to higher accuracy in both major and minor classes rather than non-ensemble classifiers, the class imbalance problem remained unsolved.

To sum up, a vast amount of effort has been put into the area of handling imbalanced mode choice datasets. However, none have reached a comprehensive and solid solution to the problem at hand, which calls for robust and innovative approaches. 

\section{Methodology}
\subsection{Ensemble Synthesizer (ENSY)}
In this section, we describe the proposed data augmentation approach designed to address the class imbalance in the dataset. The methodology combines two stages which are the generator and the validator. The generator produces synthetic instances for each class using a probabilistic approach and a classification-based validator decides whether the generated sample should be discarded or used for augmentation. The objective is to generate high-quality synthetic samples for the classes while ensuring that only instances recognized as belonging to said classes are added to the training data.

\begin{figure}[h]
  \centering
  \includegraphics[width=3.6in]{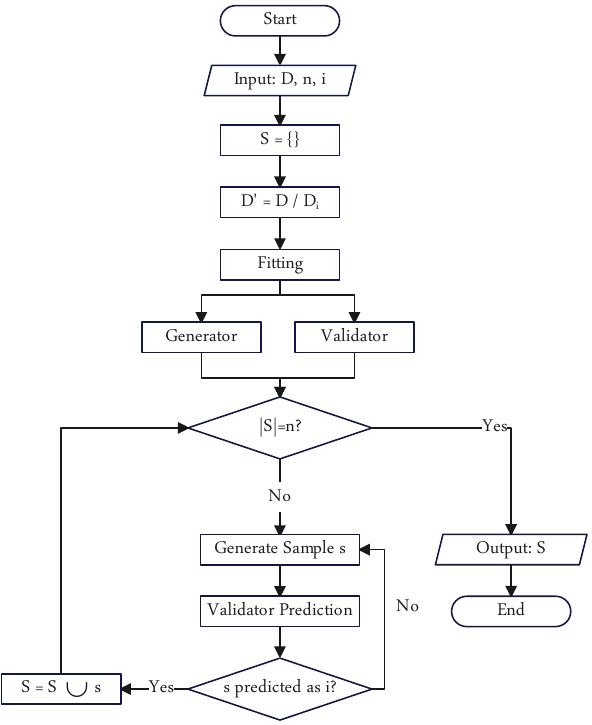}
  \caption{ENSY Flowchart}
  \label{fig:figure1}
\end{figure}

\subsubsection{Generator}
Fig.~\ref{fig:figure1} shows the flowchart of the ENSY method. As mentioned, the generator leverages probability distributions to create synthetic instances. Synthetic sample generation occurs independently for each class, and the probabilities are derived from the distribution of the remaining classes, i.e. to generate a synthetic sample for class \textit{i} in \textit{D}, instances belonging to class \textit{i} are excluded from \textit{D} and the resulting dataset $D^{\prime}$, is then passed through the generator. Furthermore, each feature is also synthesized separately nd then the generated value for each feature is concatenated to create one row of synthetic instances. 

\paragraph{Numerical Features}
To discern the underlying patterns of each numerical feature efficiently, a Gaussian Mixture Model (GMM) is fitted to the values of the feature in dataset $D^{\prime}$. A GMM is a probabilistic model that represents a mixture of several Gaussian (normal) distributions. Mathematically, a GMM is defined as follows:

For our one-dimensional case, the probability density function (PDF) of a GMM with \textit{K} components is given by:

\begin{equation}
\text{p(x)} = \sum_{i=1}^{K} \pi_i \cdot \mathcal{N}(x|\mu_i, \sigma_i^2) 
\label{eq:pdf}
\end{equation}

\begin{figure}[h]
  \centering
  \includegraphics[width=3.5in]{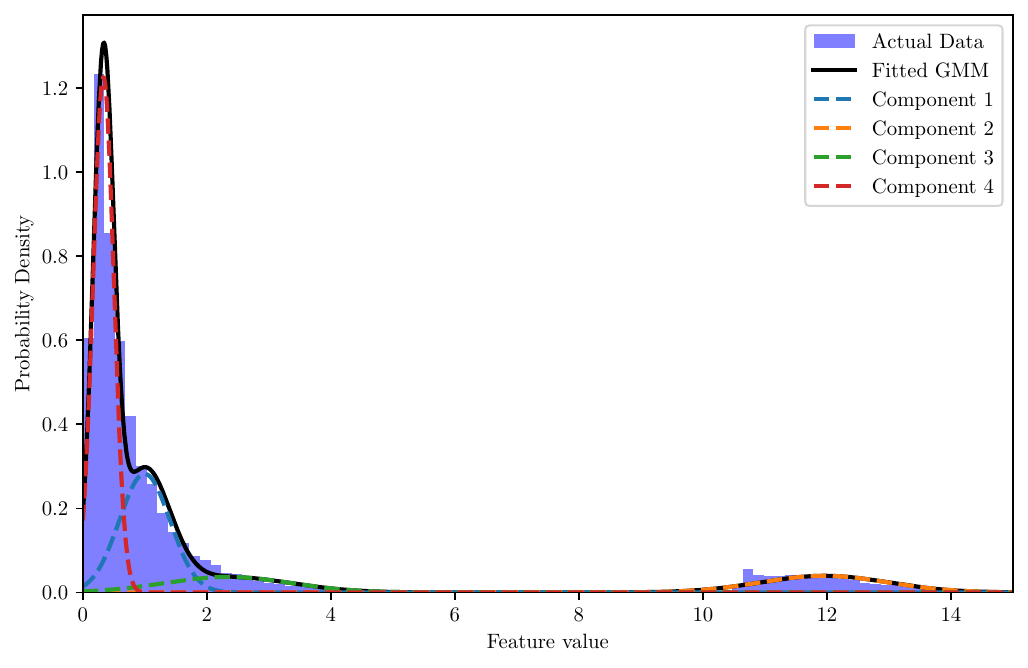}
  \caption{Actual Data, Fitted GMM, and Component Distributions}
  \label{fig:figure2}
\end{figure}

Where:
\begin{itemize}
    \item \textit{x} is the variable (feature) being modeled.
    \item $\pi_i$ is the weight of the $i$-th component, representing the probability of choosing the $i$-th Gaussian distribution. The weights $\pi_i$ are non-negative and sum to 1. They represent the contribution of each Gaussian component to the overall distribution. Larger weights mean that the corresponding Gaussian component has a more significant influence.
    \item $\mathcal{N}(x|\mu_i, \sigma_i^2)$ is the $i$-th Gaussian distribution with mean $\mu_i$ and variance $\sigma_i^2$, which is given by:
    	\begin{equation}
		N(x|\mu_i, \sigma_i^2) = \frac{1}{\sqrt{2\pi\sigma_i}} \exp\left(-\frac{(x-\mu_i)^2}{2\sigma_i^2}\right)
        \end{equation}
\end{itemize}
The process of fitting a GMM involves estimating the parameters $\pi_i$, $\mu_i$, and $\sigma_i^2$ from the data. This is often done through the Expectation-Maximization (EM) algorithm \citep{Sugiyama2016}, where the algorithm iteratively refines its estimates. Fig.~\ref{fig:figure2} depicts an example of a fitted GMM on a numerical feature.

It is important to note that while GMM efficiently captures the underlying patterns of one-dimensional numerical features, the choice of the number of components must be carefully considered to accurately represent the actual data. Additionally, during sampling using GMM, some generated values may exceed the minimum and maximum values of the actual data. This can be particularly problematic for features that logically cannot exceed a certain value (e.g., a cost feature that cannot be negative). In such cases, generated values exceeding a predefined limit can either be discarded or converted to the limit value.

\paragraph{Categorical Features}
For each categorical feature \textit{C} in the dataset $D^{\prime}$, synthetic values are generated based on random numbers between 0 and 1, mapped according to the observed frequencies of different categories. Let \textit{C} have \textit{K} unique categories. Calculate the cumulative distribution function, CDF($c_i$) for each category based on the observed frequencies in $D^{\prime}$:

\begin{equation}
CDF(c_i) = \sum_{j=1}^{i} P(c_i), \quad 1 \leq i \leq K
\end{equation}

Where $P(c_i)$ is the observed probability of category $c_i$ in $D^{\prime}$. Next, generate a random number \textit{R} between 0 and 1, then map it to a category $c_i$ based on the CDF values:

\begin{equation}
R_C = \min\{c_i : R \leq CDF(c_i)\}
\end{equation}

Where $R_C$ is the synthetic categorical value generated for categorical feature \textit{C}. This approach ensures that synthetic values for the categorical feature are generated randomly based on the observed frequencies, maintaining the distribution of the original dataset.

In essence, the generator plays a crucial role in generating synthetic instances, ensuring that the generated values for each feature mirror the distribution seen in the modified dataset, $D^{\prime}$. Operating with a straightforward yet effective approach, it iteratively processes each feature, whether categorical or numerical, to generate values aligned with the underlying patterns of $D^{\prime}$. These feature-specific values are then concatenated, resulting in a comprehensive synthetic instance.

\subsubsection{Validator}
The validator is pivotal in ensuring the quality of generated instances, ultimately enhancing classification accuracy. To achieve this, we begin by training a classifier on the original dataset. This classifier, meticulously tuned for accurate classification and adept at learning feature boundaries specific to each class, serves as a benchmark for evaluating the generated instances. Our empirical evidence supports the use of SVM or XGB, yielding superior results in this context.

Subsequently, the generated instances from the generator undergo scrutiny by the trained classifier. If an instance is identified as not belonging to class \textit{i}, it is promptly discarded. Conversely, instances correctly classified are stored for subsequent augmentation. This iterative process of generation and validation continues until the desired number of high-quality instances for class \textit{i} is achieved. By employing this meticulous validation step, we ensure that the generated instances align closely with the characteristics of class \textit{i}, contributing to a more refined and accurate augmentation process.

\subsection{Baseline Methods}
In this section, we introduce baseline data augmentation techniques chosen for their contributions to the field. These methods act as benchmarks for evaluating our model, ENSY. We explore the strengths and limitations of each baseline technique, offering reasons for the development of ENSY. Our goal is to show that ENSY effectively tackles class imbalance and improves synthetic data generation, as demonstrated through a thorough evaluation of performance metrics and model architecture.
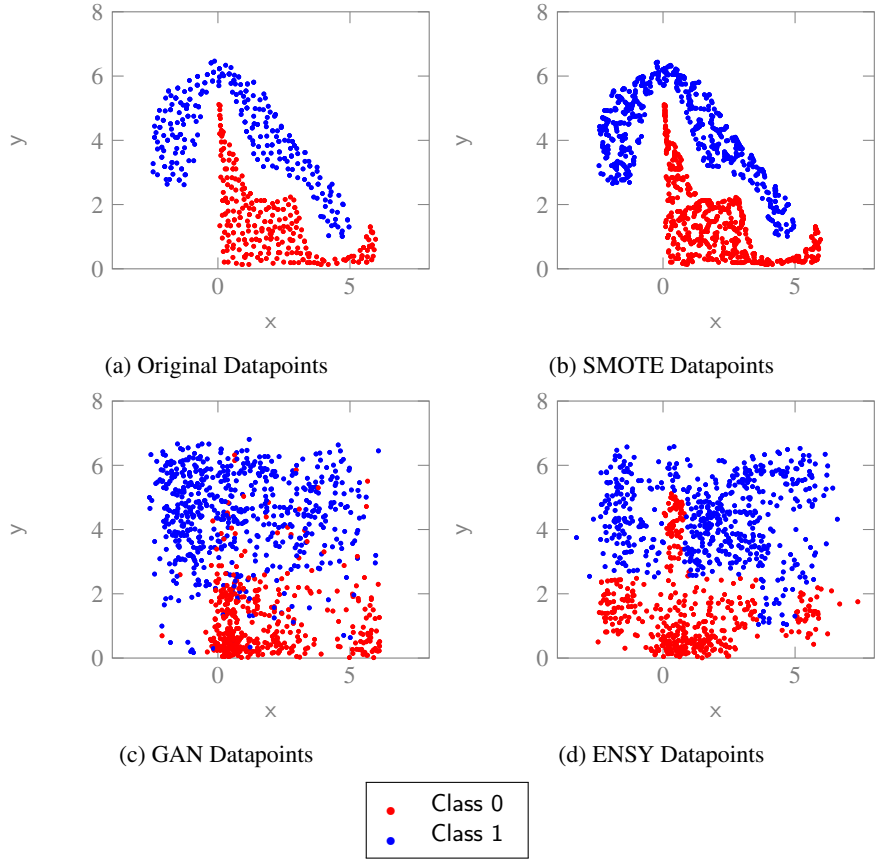
\begin{figure*}[!t]
  \centering
  \begin{subfigure}{0.35\textwidth}
    \centering
    \begin{tikzpicture}
      \begin{axis} [xmin=-4, xmax=8, xlabel={x}, ylabel={y},
        ymin=0, ymax=8, width=\textwidth, 
        scatter/classes={
          0={mark=*,red},
          1={mark=*,blue}
        }]
        \addplot[scatter, only marks, scatter src=explicit, mark options={mark size=0.7pt}] table[x=x, y=y, meta=class] {org.txt};
        \label{plot:Class0}
        \addplot[scatter, only marks, scatter src=explicit, mark options={mark size=0.7pt}] table[x=x, y=y, meta=class] {org.txt}; 
        \label{plot:Class1}
      \end{axis}
    \end{tikzpicture}
    \caption{Original Datapoints}\label{fig:figure3a}
  \end{subfigure}
  \begin{subfigure}{0.35\textwidth}
    \centering
    \begin{tikzpicture}
      \begin{axis} [xmin=-4, xmax=8, xlabel={x}, ylabel={y},
        ymin=0, ymax=8, width=\textwidth, 
        scatter/classes={
          0={mark=*,red},
          1={mark=*,blue}
        }]
        \addplot[scatter, only marks, scatter src=explicit, mark options={mark size=0.7pt}] table[x=x, y=y, meta=class] {smote.txt}; 
        \label{plot:Class0}  
        \addplot[scatter, only marks, scatter src=explicit, mark options={mark size=0.7pt}] table[x=x, y=y, meta=class] {smote.txt}; 
        \label{plot:Class1}
      \end{axis}
    \end{tikzpicture}
    \caption{SMOTE Datapoints}\label{fig:figure3b}   
  \end{subfigure}
  \begin{subfigure}{0.35\textwidth}
    \centering
    \begin{tikzpicture}
      \begin{axis} [xmin=-4, xmax=8, xlabel={x}, ylabel={y},
        ymin=0, ymax=8, width=\textwidth, 
        scatter/classes={
          0={mark=*,red},
          1={mark=*,blue}
        }]
        \addplot[scatter, only marks, scatter src=explicit, mark options={mark size=0.7pt}] table[x=x, y=y, meta=class] {gan.txt};
        \label{plot:Class0}  
        \addplot[scatter, only marks, scatter src=explicit, mark options={mark size=0.7pt}] table[x=x, y=y, meta=class] {gan.txt};
        \label{plot:Class1}
      \end{axis}
    \end{tikzpicture}
    \caption{GAN Datapoints}\label{fig:figure3c}   
  \end{subfigure}
  \begin{subfigure}{0.35\textwidth}
    \centering
    \begin{tikzpicture}
      \begin{axis} [xmin=-4, xmax=8, xlabel={x}, ylabel={y},
        ymin=0, ymax=8, width=\textwidth, 
        scatter/classes={
          0={mark=*,red},
          1={mark=*,blue}
        }]
        \addplot[scatter, only marks, scatter src=explicit, mark options={mark size=0.7pt}] table[x=x, y=y, meta=class] {ensy.txt};
        \label{plot:Class0}  
        \addplot[scatter, only marks, scatter src=explicit, mark options={mark size=0.7pt}] table[x=x, y=y, meta=class] {ensy.txt};
        \label{plot:Class1}
      \end{axis}
    \end{tikzpicture}
    \caption{ENSY Datapoints}\label{fig:figure3d}   
  \end{subfigure}
  \begin{subfigure}{\textwidth}
    \centering
    \begin{tikzpicture}
      \node[below] at ($(current bounding box.south)!0.5!(current bounding box.south |- current axis.south)$) {
        \begin{minipage}{\linewidth}
          \centering
	 \boxed{
	  \begin{tabular}{c c}
	    \textcolor{red}{\circle*{3pt}} & Class 0 \\
	    \textcolor{blue}{\circle*{3pt}} & Class 1
	  \end{tabular}
	}
        \end{minipage}
      };
    \end{tikzpicture}
  \end{subfigure}
  \caption{Generated Datapoints With Different Methods}\label{fig:combined}
\end{figure*}

\subsubsection{Random Oversampling}\label{subsubsec3}
Random Oversampling (ROS) is a commonly used method to tackle imbalances in ML datasets. It involves randomly duplicating instances from the minority class, helping balance class distributions. The simplicity of ROS is a strong point, providing an easy yet effective way to expose the classifier to more examples from the minority class.

However, ROS has its limitations. It tends to duplicate existing patterns in minority classes without adding new information. In contrast, our method, ENSY, takes a different approach. As shown in Fig.~\ref{fig:figure3d}, ENSY not only explores patterns but actively enhances the classification process by creating diverse synthetic instances. This sets ENSY apart, not just by balancing class distributions but also by introducing valuable variations. This could potentially improve the classifier's ability to recognize detailed patterns within the minority class.

\subsubsection{Synthetic Minority Over-sampling Technique for Nominal and Continuous Features (SMOTE-NC)}\label{subsubsec3}
This method is an extension of the original SMOTE algorithm, specifically designed for datasets featuring both nominal and continuous features. Aimed at addressing the class imbalance, SMOTE-NC combines oversampling and feature space interpolation to generate synthetic instances for the minority class \citep{Chawla2002}. The key steps of the SMOTE-NC algorithm are as follows:

\paragraph{Nominal Feature Handling}
For nominal features, SMOTE-NC employs a modified approach to generate synthetic instances. For each minority class instance, the algorithm identifies its \textit{k} nearest neighbors within the same class. A synthetic instance is then created by randomly selecting neighbors and replacing nominal features with the mode of the selected neighbors \citep{Chawla2002}.

\paragraph{Continuous Feature Handling}
For continuous features, SMOTE-NC follows the traditional SMOTE procedure. It selects a minority class instance and its \textit{k} nearest neighbors. A synthetic instance is generated by interpolating the continuous feature values of the selected neighbors \citep{Chawla2002}.

\paragraph{Combined Synthesis}
Synthetic instances for nominal and continuous features are integrated to form the final augmented dataset \citep{Chawla2002}.
While SMOTE-NC confines the generated data within the initial dataset's bounds through interpolation (Fig.~\ref{fig:figure3b}), ENSY takes a different path by exploring the hyperplane. A distinguishing aspect of ENSY's methodology is grounded in the understanding that validated data, drawn from the distribution of the rest of the classes, naturally exhibits a skewness towards the borderlines of that class’s data points, as opposed to its center. This skewness towards the borderlines aids the classifier in learning these crucial distinctions better, contributing to improved differentiation among the classes.

\subsubsection{Conditional Tabular Generative Adversarial Network (CTGAN)}\label{subsubsec3}
Generative Adversarial Networks (GANs) form a class of machine learning models where a generator and discriminator are pitted against each other in a training process. The generator aims to create synthetic data that is indistinguishable from real data, while the discriminator's role is to differentiate between genuine and synthetic samples. This adversarial setup assists in the improvement of both the generator's ability to create realistic data and the discriminator's sharpness in distinguishing between real and generated instances.

In our exploration of data augmentation techniques, we delve into Conditional Tabular GAN (CTGAN) \citep{Xu2019}. This specialized GAN variant is designed to synthesize tabular data, particularly effective for datasets with a mix of categorical and numerical features. It leverages conditional generative modeling, allowing us to control the characteristics of the generated data, making it well-suited for applications where preserving specific attributes is crucial. Furthermore, CTGAN can be conditioned on the minority class label, enabling the targeted generation of synthetic instances for the underrepresented class.

CTGAN consists of a generator network and a discriminator (or critic) network. The generator network is trained to minimize generator loss, encouraging the generation of synthetic samples that are difficult for the discriminator to distinguish from real instances. Conversely, the discriminator aims to minimize discriminator loss, distinguishing between real and synthetic samples accurately.

\begin{figure}[h]
  \centering
  \begin{tikzpicture}
    \begin{axis} [name=plot, xmin=0, xmax=500, xlabel={epoch}, ylabel={loss},
      ymin=-3.5, ymax=2, width=.5\textwidth]
      \addplot[red, mark=o, mark size=0.5pt] table{dis.txt}; \label{plot:discriminator_loss}
      \addplot[blue, mark=o, mark size=0.5pt] table{gen.txt}; \label{plot:generator_loss}
    \end{axis}
    \node[anchor=north east, draw=black, fill=white] (legend) at (plot.north east) {
      \begin{tabular}{c|c}
        Discriminator Loss & \ref{plot:discriminator_loss} \\
        Generator Loss & \ref{plot:generator_loss}
      \end{tabular}
    };
  \end{tikzpicture}
  \caption{Generator and Discriminator Loss Over 500 Epochs in CTGAN}
  \label{fig:figure4}
\end{figure}
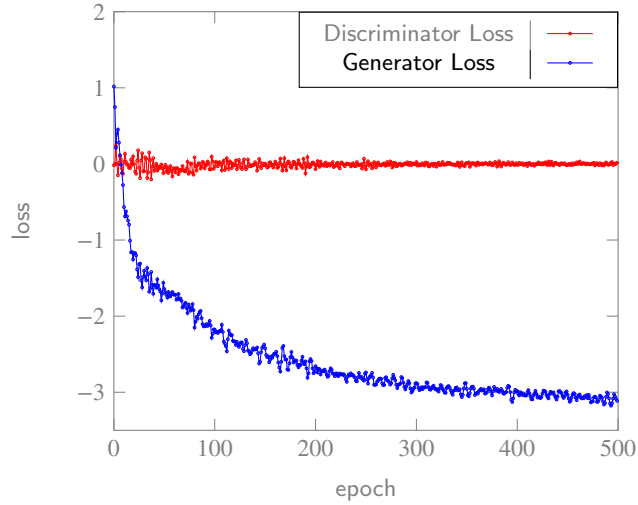

Despite CTGAN's efficacy in generating synthetic values, achieving optimal training requires careful tuning for a smooth and reliable process. While Python packages such as SDV and YData provide extensive modification options for CTGAN implementation, our approach employs an open-source version for more nuanced monitoring, enabling precise adjustments at every stage of the model. Fig.~\ref{fig:figure4} illustrates the training progress of our implementation.

The objective during training is for the discriminator loss to oscillate around 0, indicating its inability to discern between real and fake samples, while the generator loss should stabilize at a negative value, signifying the realism of the generated data successfully fooling the discriminator.

Despite its significant capabilities, CTGAN is not without its shortcomings. When the model is trained on the entire dataset, it tends to overfit the majority classes, neglecting the intricate patterns within the minority classes. Conversely, fitting the model on each class independently poses challenges as it tends to generate noise due to its limited understanding of the patterns present in the rest of the classes (Fig.~\ref{fig:figure3c}). Furthermore, as the dataset size increases, the computational demands become more time-consuming, and fine-tuning becomes a challenging task that demands extensive knowledge and trial-and-error exploration.

\subsection{Metrics}\label{subsec3}
In this study, we employ several key metrics to evaluate the performance of data augmentation techniques. These metrics provide insights into different aspects of classification accuracy.

\subsubsection{Overall Accuracy}\label{subsubsec3}
Overall accuracy is a commonly used metric to assess the general correctness of a classification model \citep{Vujovic2021}, \citep{Hossin2015}. It is calculated as the ratio of correctly predicted instances to the total number of instances in the dataset.

\begin{equation}
\text{Overall Accuracy} = \frac{\text{Number of Correct Predictions}}{\text{Total Number of Instances}}
\label{eq:overall_accuracy}
\end{equation}

\subsubsection{Precision}\label{subsubsec3}
Precision measures the accuracy of positive predictions made by the model. It is calculated as the ratio of correctly predicted positive instances to the total number of positive predictions (true positives and false positives).

\begin{equation}
\text{Precision} = \frac{\text{True Positives}}{\text{True Positives + False Positives}}
\label{eq:precision}
\end{equation}

\subsubsection{Recall (Sensitivity)}\label{subsubsec3}
Recall, also known as sensitivity or true positive rate, evaluates the model's ability to capture all positive instances. It is calculated as the ratio of correctly predicted positive instances to the total number of actual positive instances (true positives and false negatives).

\begin{equation}
\text{Recall} = \frac{\text{True Positives}}{\text{True Positives + False Negatives}}
\label{eq:recall}
\end{equation}

\subsubsection{F1-score}\label{subsubsec3}
The F1-score provides a balance between precision and recall. In other words, it corresponds to the harmonic average of the precision and the recall. This metric is especially useful in imbalanced datasets.

\begin{equation}
\text{F1} = \frac{2 \times (\text{Precision} \times \text{Recall})}{\text{Precision} + \text{Recall}}
\label{eq:F1}
\end{equation}

These metrics collectively provide a comprehensive assessment of the classification performance, considering aspects such as overall correctness, precision in positive predictions, and the ability to identify all positive instances.

\section{Dataset Description}\label{sec4}
The London Passenger Mode Choice (LPMC) \citep{Hillel2018} dataset, derived from the London Travel Demand Survey (LTDS), comprises 81,086 trips made by 31,954 individuals across 17,616 households over three years (April 2012 to March 2015). On the other hand, the Korea Transport Database \citep{KTDB2016} is based on the national household travel survey data collected in South Korea in 2016. The survey was conducted in 202,316 households. These two datasets serve as valuable resources for understanding urban multi-modal transport network dynamics and predicting mode choice (Table~\ref{tab:table1}, ~\ref{tab:table2}). 

\begin{table}[h]
\caption{Mode Shares in LPMC\label{tab:table1}}
\begin{tabular}{lccl}
\hline
Travel Mode & Number of Trips & Mode Share (\%) \\
\hline
Walking & 14268 &17.596 \\
Cycling & 2405 & 2.966 \\
Public transport & 28605 & 35.277 \\
Driving & 35808 & 44.161 \\
Total & 81086 & 100.00 \\
\hline
\end{tabular}
\end{table}

\begin{table}[h]
\caption{Mode Shares in KTDB\label{tab:table2}}
\begin{tabular}{lccl}
\hline
Travel Mode & Number of Trips & Mode Share (\%) \\
\hline
Auto & 33132 & 16.850 \\
Riding & 6127 & 3.116 \\
Subway & 21676 & 11.024 \\
Bus & 37155 & 18.896 \\
Subway and Bus & 9855 & 5.012 \\
Taxi & 2009 & 1.022 \\
Cycling & 3432 & 1.745 \\
Walk & 83245 & 42.335 \\
total & 196631 & 100.00 \\
\hline
\end{tabular}
\end{table}

The raw datasets need to undergo various preprocessing steps before they can be analyzed further. Firstly, columns that do not add valuable information, such as unique identifiers, are eliminated from the dataset. Next, features with strong linear correlations to other features are excluded to guarantee that the remaining features possess valuable and unique information. The last step involves generating new features derived from existing columns to eliminate redundancy and simplify the dataset. As these datasets do not contain any duplicate rows or missing values, the modified versions do not require any additional modifications. In this paper, the modified LPMC dataset comprises 17 features, including 8 categorical (Trip purpose, Car ownership, ...) and 9 numerical (Distance, Age, Time, ...) features. Conversely, the modified KTDB contains 17 features, with 12 being categorical (Age Range, Trip purpose, Income Range, ...) and 5 being numerical (Time, Distance, Cost, ...).

\section{Results and Discussion}\label{sec5}
In this section, we present the outcomes of our experiments, comparing the classification performance of baseline methods (ROS, SMOTE-NC, CTGAN) with our proposed method, ENSY. The experiments were conducted on modified versions of the LPMC and KTDB datasets, randomly split into training and test datasets, with 15\% allocated to the test set.

As evidenced by the results presented in Table~\ref{tab:table3} and Table~\ref{tab:table4}, ENSY consistently showcased notable improvements in the F1-score for minority classes across diverse scenarios. Given that the F1-score strikes a balance between precision and recall, these findings suggest that ENSY effectively addresses class imbalance.

Analyzing the performance of different classification models, we observed that XGB consistently achieved the best results, followed by Random Forest, while Neural Networks (NN) yielded comparatively lower performance.

Surprisingly, data augmentation, including ENSY, did not significantly enhance classification performance when Random Forest was employed. In these scenarios, the raw dataset outperformed the augmented counterparts, aligning with findings from \citep{Chen2023}.

An interesting observation emerged when the entire dataset was augmented using CTGAN before splitting. In this case, RF, XGB, and NN consistently achieved overall accuracies exceeding 99\%, aligning with \citep{Majeed2023}. However, this trend did not hold when only the training dataset underwent augmentation.

Table~\ref{tab:table5} and Table~\ref{tab:table6} illustrate that RF achieves the highest overall accuracy in raw data. However, ESNY demonstrates significant potential for enhancing overall accuracy in both XGB and NN. Once more, the tables highlight that XGB outperforms RF and NN in terms of performance.

These findings underscore the nuanced impact of data augmentation techniques on different classification models. While ENSY showcased consistent improvements for minority classes, the choice of classifier and the nature of augmentation can significantly influence overall performance.

\begin{table*}[!t]
\caption{Classification Reports-LMPC}
\centering
\resizebox{\linewidth}{!}{
	\begin{tabular}{ccccccccccccccccccc}
	\toprule
	\textbf{Classifier} & \textbf{Minority Class} & \multicolumn{5}{c}{\textbf{Precision}} & \multicolumn{5}{c}{\textbf{Recall}} & \multicolumn{5}{c}	{\textbf{F1-score}} \\
	\cmidrule(lr){3-7} \cmidrule(lr){8-12} \cmidrule(lr){13-17} 
	& \textbf{} & \textbf{Raw} & \textbf{ROS} & \textbf{SMOTE-NC} & \textbf{CTGAN} & \textbf{ENSY} & \textbf{Raw} & \textbf{ROS} & \textbf{SMOTE-NC} & \textbf{CTGAN} & \textbf{ENSY} & \textbf{Raw} & \textbf{ROS} & \textbf{SMOTE-NC} & \textbf{CTGAN} & \textbf{ENSY} \\
	\midrule
	\multirow{2}{*}{\textbf{XGB}} & Walking & 0.78 & 0.76 & 0.73 & 0.78 & \textbf{0.80} & 0.75 & 0.77 & \textbf{0.80} & 0.75 & 0.79 & 0.77 & 0.77 & 0.76 & 0.77 & \textbf{0.80} \\
	& Cycling & 0.64 & 0.56 & 0.35 & 0.52 & \textbf{0.67} & 0.07 & 0.14 & 0.21 & 0.07 & \textbf{0.37} & 0.12 & 0.23 & 0.26 & 0.12 & \textbf{0.47} \\
	\midrule
	\multirow{2}{*}{\textbf{RF}} & Walking & \textbf{0.77} & 0.73 & 0.68 & \textbf{0.77} & 0.75 & 0.74 & 0.78 & \textbf{0.81} & 0.75 & 0.75 & \textbf{0.76} & \textbf{0.76} & 0.74 & \textbf{0.76} & 0.75 \\
	& Cycling & \textbf{0.96} & 0.76 & 0.33 & 0.68 & 0.51 & 0.05 & 0.10 & 0.21 & 0.05 & \textbf{0.27} & 0.09 & 0.18 & 0.25 & 0.09 & \textbf{0.35} \\
	\midrule
	\multirow{2}{*}{\textbf{NN}} & Walking & 0.69 & 0.70 & \textbf{0.71} & 0.66 & \textbf{0.71} & 0.67 & 0.66 & 0.65 & \textbf{0.70} & 0.66 & \textbf{0.68} & \textbf{0.68} & \textbf{0.68} & \textbf{0.68} & \textbf{0.68} \\
	& Cycling & 0.27 & \textbf{0.44} & 0.27 & 0.18 & 0.37 & 0.01 & 0.02 & 0.01 & 0.20 & \textbf{0.32} & 0.01 & 0.03 & 0.02 & 0.19 & \textbf{0.34} \\
	\bottomrule
	\end{tabular}
}
\label{tab:table3}
\end{table*}

\begin{table*}[!t]
\caption{Classification Reports-KTDB}
\centering
\resizebox{\linewidth}{!}{
	\begin{tabular}{ccccccccccccccccccc}
	\toprule
	\textbf{Classifier} & \textbf{Minority Class} & \multicolumn{5}{c}{\textbf{Precision}} & \multicolumn{5}{c}{\textbf{Recall}} & \multicolumn{5}{c}	{\textbf{F1-score}} \\
	\cmidrule(lr){3-7} \cmidrule(lr){8-12} \cmidrule(lr){13-17} 
	& \textbf{} & \textbf{Raw} & \textbf{ROS} & \textbf{SMOTE-NC} & \textbf{CTGAN} & \textbf{ENSY} & \textbf{Raw} & \textbf{ROS} & \textbf{SMOTE-NC} & \textbf{CTGAN} & \textbf{ENSY} & \textbf{Raw} & \textbf{ROS} & \textbf{SMOTE-NC} & \textbf{CTGAN} & \textbf{ENSY} \\
	\midrule
	\multirow{2}{*}{\textbf{XGB}} & Taxi & 0.41 & 0.31 & 0.16 & 0.34 & \textbf{0.47} & 0.13 & 0.16 & 0.21 & 0.15 & \textbf{0.33} & 0.20 & 0.21 & 0.19 & 0.21 & \textbf{0.38} \\
	& Cycling & 0.69 & 0.62 & 0.46 & 0.66 & \textbf{0.71} & 0.49 & 0.53 & \textbf{0.55} & 0.50 & \textbf{0.55} & 0.58 & 0.57 & 0.50 & 0.57 & \textbf{0.62} \\
	\midrule
	\multirow{2}{*}{\textbf{RF}} & Taxi & \textbf{0.39} & 0.27 & 0.12 & 0.13 & 0.31 & 0.07 & 0.09 & 0.17 & 0.09 & \textbf{0.22} & 0.11 & 0.13 & 0.14 & 0.11 & \textbf{0.26} \\
	& Cycling & \textbf{0.78} & 0.69 & 0.40 & 0.63 & 0.66 & 0.43 & 0.48 & \textbf{0.54} & 0.45 & 0.41 & \textbf{0.56} & \textbf{0.56} & 0.46 & 0.53 & 0.50 \\
	\midrule
	\multirow{2}{*}{\textbf{NN}} & Taxi & 0.22 & 0.13 & 0.17 & 0.07 & \textbf{0.32} & 0.11 & 0.15 & 0.11 & 0.10 & \textbf{0.17} & 0.15 & 0.14 & 0.14 & 0.08 & \textbf{0.22} \\
	& Cycling & 0.62 & 0.39 & 0.43 & 0.42 & \textbf{0.70} & 0.43 & \textbf{0.52} & 0.47 & 0.43 & 0.46 & 0.50 & 0.44 & 0.45 & 0.42 & \textbf{0.55} \\
	\bottomrule
	\end{tabular}
}
\label{tab:table4}
\end{table*}

\begin{table}[h]
\caption{Overall Accuracy (\%)-LPMC }
\label{tab:acc}
\centering
\begin{tabular}{ccccccc}
\hline
\textbf{Method} & \textbf{Raw} & \textbf{ROS} & \textbf{SMOTE-NC} & \textbf{CTGAN} & \textbf{ENSY} \\
\hline
XGB & 80.84 & 81.04 & 80.42 & 81.21 & \textbf{83.23} \\
RF & \textbf{80.36} & 80.04 & 78.66 & 80.34 & 80.04 \\
NN & 75.41 & 75.42 & \textbf{75.55} & 74.07 & 75.46 \\
\hline
\end{tabular}
\label{tab:table5}
\end{table}

\begin{table}[h]
\caption{Overall Accuracy (\%)-KTDB }
\label{tab:acc}
\centering
\begin{tabular}{ccccccc}
\hline
\textbf{Method} & \textbf{Raw} & \textbf{ROS} & \textbf{SMOTE-NC} & \textbf{CTGAN} & \textbf{ENSY} \\
\hline
XGB & 72.46 & 71.81 & 69.95 & 71.74 & \textbf{73.51} \\
RF & \textbf{71.32} & 70.46 & 67.89 & 69.42 & 70.83 \\
NN & 67.47 & 62.89 & 65.49 & 64.32 & \textbf{69.32} \\
\hline
\end{tabular}
\label{tab:table6}
\end{table}

\section{Conclusion}\label{sec6}
In conclusion, our study underscores the success of ENSY in mitigating class imbalance and enhancing overall accuracy, as evidenced by a notable 3\% improvement in overall accuracy for the LPMC dataset. Furthermore, the almost quadrupled F-1 score for the minority class (Cycling) reflects a significant advancement in addressing the challenges posed by imbalanced datasets.

Similarly, for the KTDB dataset, ENSY demonstrated its effectiveness by contributing to an improvement of almost 1.5\% in overall accuracy, coupled with an impressive doubling of the F-1 score for the minority class (Taxi). These results affirm the potential of ENSY as a valuable tool in mode choice prediction models, showcasing its ability to offer substantial performance gains in different contexts.

Notably, the improvements were particularly pronounced when employing XGB, which consistently outperformed Neural Networks (NN) and Random Forest (RF), even in scenarios without augmentation. This highlights the robustness and efficacy of ENSY, especially in conjunction with XGB. The effects of different models used as validators for ENSY, such as NN, warrant further exploration and study to understand their nuances and potential impact on model performance.

While the improved accuracy and F-1 scores are promising, they prompt a necessary reflection on whether the efforts invested in developing and implementing ENSY are justified. Our study suggests that, despite the advancements, there exist trade-offs that necessitate ongoing research and refinement. Exploring avenues for further improvement remains an essential aspect of future work, with potential areas of exploration including parameter tuning, ensemble methods, and additional feature engineering.

In summary, our research positions ENSY as a promising solution to the challenges of class imbalance in mode choice prediction models. However, ongoing efforts are crucial to optimize its performance, understand its limitations, and justify its application in practical transportation scenarios.

\bibliographystyle{cas-model2-names}

\bibliography{reference}

\begin{thebibliography}{37}
\expandafter\ifx\csname natexlab\endcsname\relax\def\natexlab#1{#1}\fi
\providecommand{\url}[1]{\texttt{#1}}
\providecommand{\href}[2]{#2}
\providecommand{\path}[1]{#1}
\providecommand{\DOIprefix}{doi:}
\providecommand{\ArXivprefix}{arXiv:}
\providecommand{\URLprefix}{URL: }
\providecommand{\Pubmedprefix}{pmid:}
\providecommand{\doi}[1]{\href{http://dx.doi.org/#1}{\path{#1}}}
\providecommand{\Pubmed}[1]{\href{pmid:#1}{\path{#1}}}
\providecommand{\bibinfo}[2]{#2}
\ifx\xfnm\relax \def\xfnm[#1]{\unskip,\space#1}\fi
\bibitem[{Aziira et~al.(2020)Aziira, Setiawan and Soesanti}]{Aziira2020}
\bibinfo{author}{Aziira, A.H.}, \bibinfo{author}{Setiawan, N.A.},
  \bibinfo{author}{Soesanti, I.}, \bibinfo{year}{2020}.
\newblock \bibinfo{title}{Generation of synthetic continuous numerical data
  using generative adversarial networks}.
\newblock \bibinfo{journal}{J. Phys. Conf. Ser.} \bibinfo{volume}{1577},
  \bibinfo{pages}{012027}.
\bibitem[{Batuwita and Palade(2010)}]{Batuwita2010}
\bibinfo{author}{Batuwita, R.}, \bibinfo{author}{Palade, V.},
  \bibinfo{year}{2010}.
\newblock \bibinfo{title}{{FSVM-CIL}: Fuzzy support vector machines for class
  imbalance learning}.
\newblock \bibinfo{journal}{IEEE Trans. Fuzzy Syst.} \bibinfo{volume}{18},
  \bibinfo{pages}{558--571}.
\bibitem[{Birr(2018)}]{Birr2018}
\bibinfo{author}{Birr, K.}, \bibinfo{year}{2018}.
\newblock \bibinfo{title}{Mode choice modelling for urban areas}.
\newblock \bibinfo{journal}{Czas. Tech.} \bibinfo{volume}{6}.
\bibitem[{Chaipanha and Kaewwichian(2022)}]{Chaipanha2022}
\bibinfo{author}{Chaipanha, W.}, \bibinfo{author}{Kaewwichian, P.},
  \bibinfo{year}{2022}.
\newblock \bibinfo{title}{Smote vs. random undersampling for imbalanced data -
  car ownership demand model}.
\newblock \bibinfo{journal}{Komunik{\'a}cie} \bibinfo{volume}{24},
  \bibinfo{pages}{D105--D115}.
\bibitem[{Chawla et~al.(2002)Chawla, Bowyer, Hall and Kegelmeyer}]{Chawla2002}
\bibinfo{author}{Chawla, N.V.}, \bibinfo{author}{Bowyer, K.W.},
  \bibinfo{author}{Hall, L.O.}, \bibinfo{author}{Kegelmeyer, W.P.},
  \bibinfo{year}{2002}.
\newblock \bibinfo{title}{{SMOTE}: Synthetic minority over-sampling technique}.
\newblock \bibinfo{journal}{J. Artif. Intell. Res.} \bibinfo{volume}{16},
  \bibinfo{pages}{321--357}.
\bibitem[{Chen and Cheng(2023)}]{Chen2023}
\bibinfo{author}{Chen, H.}, \bibinfo{author}{Cheng, Y.}, \bibinfo{year}{2023}.
\newblock \bibinfo{title}{Travel mode choice prediction using imbalanced
  machine learning}.
\newblock \bibinfo{journal}{IEEE Trans. Intell. Transp. Syst.}
  \bibinfo{volume}{24}, \bibinfo{pages}{3795--3808}.
\bibitem[{De~Ortuzar and Willumsen(2011)}]{De_Ortuzar2011}
\bibinfo{author}{De~Ortuzar, J.}, \bibinfo{author}{Willumsen, L.G.},
  \bibinfo{year}{2011}.
\newblock \bibinfo{title}{Modelling Transport}.
\newblock \bibinfo{edition}{4} ed., \bibinfo{publisher}{Wiley}.
\bibitem[{Diallo et~al.(2022)Diallo, Lozenguez, Doniec and
  Mandiau}]{Diallo2022}
\bibinfo{author}{Diallo, A.O.}, \bibinfo{author}{Lozenguez, G.},
  \bibinfo{author}{Doniec, A.}, \bibinfo{author}{Mandiau, R.},
  \bibinfo{year}{2022}.
\newblock \bibinfo{title}{Estimation of minority modes of transportation based
  on machine learning approach}.
\newblock \bibinfo{journal}{Procedia Comput. Sci.} \bibinfo{volume}{201},
  \bibinfo{pages}{265--272}.
\bibitem[{Edward(2003)}]{Edward2003}
\bibinfo{author}{Edward, Wu, G.A.}, \bibinfo{year}{2003}.
\newblock \bibinfo{title}{{Class-Boundary} alignment for imbalanced dataset
  learning}.
\newblock \bibinfo{journal}{ICML 2003 Workshop on Learning from Imbalanced Data
  Sets} .
\bibitem[{Hensher et~al.(2015)Hensher, Rose and Greene}]{Hensher2015}
\bibinfo{author}{Hensher, D.A.}, \bibinfo{author}{Rose, J.M.},
  \bibinfo{author}{Greene, W.H.}, \bibinfo{year}{2015}.
\newblock \bibinfo{title}{Applied Choice Analysis}.
\newblock \bibinfo{publisher}{Cambridge University Press}.
\bibitem[{Hillel et~al.(2021)Hillel, Bierlaire, Elshafie and Jin}]{Hillel2021}
\bibinfo{author}{Hillel, T.}, \bibinfo{author}{Bierlaire, M.},
  \bibinfo{author}{Elshafie, M.Z.E.B.}, \bibinfo{author}{Jin, Y.},
  \bibinfo{year}{2021}.
\newblock \bibinfo{title}{A systematic review of machine learning
  classification methodologies for modelling passenger mode choice}.
\newblock \bibinfo{journal}{J. Choice Model.} \bibinfo{volume}{38},
  \bibinfo{pages}{100221}.
\bibitem[{Hillel et~al.(2018)Hillel, Elshafie and Jin}]{Hillel2018}
\bibinfo{author}{Hillel, T.}, \bibinfo{author}{Elshafie, M.Z.E.B.},
  \bibinfo{author}{Jin, Y.}, \bibinfo{year}{2018}.
\newblock \bibinfo{title}{Recreating passenger mode choice-sets for transport
  simulation: A case study of london, {UK}}.
\newblock \bibinfo{journal}{Proc. Inst. Civ. Eng. - Smart Infrastruct. Constr.}
  \bibinfo{volume}{171}, \bibinfo{pages}{29--42}.
\bibitem[{{Hossin} and {Sulaiman}(2015)}]{Hossin2015}
\bibinfo{author}{{Hossin}}, \bibinfo{author}{{Sulaiman}}, \bibinfo{year}{2015}.
\newblock \bibinfo{title}{A review on evaluation metrics for data
  classification evaluations}.
\newblock \bibinfo{journal}{Int. J. Data Min. Knowl. Manag. Process}
  \bibinfo{volume}{5}, \bibinfo{pages}{01--11}.
\bibitem[{Kadiyali(2013)}]{Kadiyali2013}
\bibinfo{author}{Kadiyali, L.R.}, \bibinfo{year}{2013}.
\newblock \bibinfo{title}{Traffic engineering and transport planning}.
\newblock \bibinfo{publisher}{Khanna Publishers}.
\bibitem[{Kim(2021)}]{Kim2021}
\bibinfo{author}{Kim, E.J.}, \bibinfo{year}{2021}.
\newblock \bibinfo{title}{Analysis of travel mode choice in seoul using an
  interpretable machine learning approach}.
\newblock \bibinfo{journal}{J. Adv. Transp.} \bibinfo{volume}{2021},
  \bibinfo{pages}{1--13}.
\bibitem[{KTDB(2016)}]{KTDB2016}
\bibinfo{author}{KTDB}, \bibinfo{year}{2016}.
\newblock \bibinfo{title}{National transport surveys}.
\newblock
  \bibinfo{howpublished}{\url{https://www.ktdb.go.kr/eng/contents.do?key=263}}.
\bibitem[{Li et~al.(2021)Li, Wang, Wu, Chen and Zhou}]{Li2021}
\bibinfo{author}{Li, X.}, \bibinfo{author}{Wang, Y.}, \bibinfo{author}{Wu, Y.},
  \bibinfo{author}{Chen, J.}, \bibinfo{author}{Zhou, J.}, \bibinfo{year}{2021}.
\newblock \bibinfo{title}{Modeling intercity travel mode choice with data
  balance changes: A comparative analysis of bayesian logit model and
  artificial neural networks}.
\newblock \bibinfo{journal}{J. Adv. Transp.} \bibinfo{volume}{2021},
  \bibinfo{pages}{1--22}.
\bibitem[{Majeed and Hwang(2023)}]{Majeed2023}
\bibinfo{author}{Majeed, A.}, \bibinfo{author}{Hwang, S.O.},
  \bibinfo{year}{2023}.
\newblock \bibinfo{title}{{CTGAN-MOS}: Conditional generative adversarial
  network based minority-class-augmented oversampling scheme for imbalanced
  problems}.
\newblock \bibinfo{journal}{IEEE Access} \bibinfo{volume}{11},
  \bibinfo{pages}{85878--85899}.
\bibitem[{McFadden(1974)}]{McFadden1974}
\bibinfo{author}{McFadden, D.}, \bibinfo{year}{1974}.
\newblock \bibinfo{title}{The measurement of urban travel demand}.
\newblock \bibinfo{journal}{J. Public Econ.} \bibinfo{volume}{3},
  \bibinfo{pages}{303--328}.
\bibitem[{Menardi and Torelli(2014)}]{Menardi2014}
\bibinfo{author}{Menardi, G.}, \bibinfo{author}{Torelli, N.},
  \bibinfo{year}{2014}.
\newblock \bibinfo{title}{Training and assessing classification rules with
  imbalanced data}.
\newblock \bibinfo{journal}{Data Min. Knowl. Discov.} \bibinfo{volume}{28},
  \bibinfo{pages}{92--122}.
\bibitem[{Mohammed et~al.(2020)Mohammed, Rawashdeh and Abdullah}]{Mohammed2020}
\bibinfo{author}{Mohammed, R.}, \bibinfo{author}{Rawashdeh, J.},
  \bibinfo{author}{Abdullah, M.}, \bibinfo{year}{2020}.
\newblock \bibinfo{title}{Machine learning with oversampling and undersampling
  techniques: Overview study and experimental results}, in:
  \bibinfo{booktitle}{2020 11th International Conference on Information and
  Communication Systems ({ICICS})}, \bibinfo{publisher}{IEEE}.
\bibitem[{Pulugurta et~al.(2013)Pulugurta, Arun and Errampalli}]{Pulugurta2013}
\bibinfo{author}{Pulugurta, S.}, \bibinfo{author}{Arun, A.},
  \bibinfo{author}{Errampalli, M.}, \bibinfo{year}{2013}.
\newblock \bibinfo{title}{Use of artificial intelligence for mode choice
  analysis and comparison with traditional multinomial logit model}.
\newblock \bibinfo{journal}{Procedia Soc. Behav. Sci.} \bibinfo{volume}{104},
  \bibinfo{pages}{583--592}.
\bibitem[{Qian et~al.(2021)Qian, Aghaabbasi, Ali, Alqurashi, Salah, Zainol,
  Moeinaddini and Hussein}]{Qian2021}
\bibinfo{author}{Qian, Y.}, \bibinfo{author}{Aghaabbasi, M.},
  \bibinfo{author}{Ali, M.}, \bibinfo{author}{Alqurashi, M.},
  \bibinfo{author}{Salah, B.}, \bibinfo{author}{Zainol, R.},
  \bibinfo{author}{Moeinaddini, M.}, \bibinfo{author}{Hussein, E.E.},
  \bibinfo{year}{2021}.
\newblock \bibinfo{title}{Classification of imbalanced travel mode choice to
  work data using adjustable {SVM} model}.
\newblock \bibinfo{journal}{Appl. Sci. (Basel)} \bibinfo{volume}{11},
  \bibinfo{pages}{11916}.
\bibitem[{Rasouli and Timmermans(2014)}]{Rasouli2014}
\bibinfo{author}{Rasouli, S.}, \bibinfo{author}{Timmermans, H.J.P.},
  \bibinfo{year}{2014}.
\newblock \bibinfo{title}{Using ensembles of decision trees to predict
  transport mode choice decisions: Effects on predictive success and
  uncertainty estimates}.
\newblock \bibinfo{journal}{Eur. J. Transp. Infrastruct. Res.} .
\bibitem[{Rezaei et~al.(2022)Rezaei, Khojandi, Haque, Brakewood, Jin and
  Cherry}]{Rezaei2022}
\bibinfo{author}{Rezaei, S.}, \bibinfo{author}{Khojandi, A.},
  \bibinfo{author}{Haque, A.M.}, \bibinfo{author}{Brakewood, C.},
  \bibinfo{author}{Jin, M.}, \bibinfo{author}{Cherry, C.},
  \bibinfo{year}{2022}.
\newblock \bibinfo{title}{Performance evaluation of mode choice models under
  balanced and imbalanced data assumptions}.
\newblock \bibinfo{journal}{Transp. Lett.} \bibinfo{volume}{14},
  \bibinfo{pages}{920--932}.
\bibitem[{Richards and Zill(2019)}]{Richards2019}
\bibinfo{author}{Richards, M.J.}, \bibinfo{author}{Zill, J.C.},
  \bibinfo{year}{2019}.
\newblock \bibinfo{title}{Modelling mode choice with machine learning
  algorithms}, \bibinfo{publisher}{Australasian Transport Research Forum
  (ATRF)}, \bibinfo{address}{Australia}.
\bibitem[{Sekhar et~al.(2016)Sekhar, {Minal} and Madhu}]{Sekhar2016}
\bibinfo{author}{Sekhar, C.R.}, \bibinfo{author}{{Minal}},
  \bibinfo{author}{Madhu, E.}, \bibinfo{year}{2016}.
\newblock \bibinfo{title}{Mode choice analysis using random forrest decision
  trees}.
\newblock \bibinfo{journal}{Transp. Res. Procedia} \bibinfo{volume}{17},
  \bibinfo{pages}{644--652}.
\bibitem[{Song et~al.(2023)Song, Wang, Ye, Zaretzki and Liu}]{Song2023}
\bibinfo{author}{Song, Y.}, \bibinfo{author}{Wang, Y.}, \bibinfo{author}{Ye,
  X.}, \bibinfo{author}{Zaretzki, R.}, \bibinfo{author}{Liu, C.},
  \bibinfo{year}{2023}.
\newblock \bibinfo{title}{Loan default prediction using a credit
  rating-specific and multi-objective ensemble learning scheme}.
\newblock \bibinfo{journal}{Inf. Sci. (Ny)} \bibinfo{volume}{629},
  \bibinfo{pages}{599--617}.
\bibitem[{Sugiyama(2016)}]{Sugiyama2016}
\bibinfo{author}{Sugiyama, M.}, \bibinfo{year}{2016}.
\newblock \bibinfo{title}{Maximum likelihood estimation for gaussian mixture
  model}, in: \bibinfo{booktitle}{Introduction to Statistical Machine
  Learning}. \bibinfo{publisher}{Elsevier}, pp. \bibinfo{pages}{157--168}.
\bibitem[{Vujovic(2021)}]{Vujovic2021}
\bibinfo{author}{Vujovic, Z.}, \bibinfo{year}{2021}.
\newblock \bibinfo{title}{Classification model evaluation metrics}.
\newblock \bibinfo{journal}{Int. J. Adv. Comput. Sci. Appl.}
  \bibinfo{volume}{12}.
\bibitem[{Wang and Japkowicz(2010)}]{Wang2010}
\bibinfo{author}{Wang, B.X.}, \bibinfo{author}{Japkowicz, N.},
  \bibinfo{year}{2010}.
\newblock \bibinfo{title}{Boosting support vector machines for imbalanced data
  sets}.
\newblock \bibinfo{journal}{Knowl. Inf. Syst.} \bibinfo{volume}{25},
  \bibinfo{pages}{1--20}.
\bibitem[{Wang and Ross(2018)}]{Wang2018}
\bibinfo{author}{Wang, F.}, \bibinfo{author}{Ross, C.L.}, \bibinfo{year}{2018}.
\newblock \bibinfo{title}{Machine learning travel mode choices: Comparing the
  performance of an extreme gradient boosting model with a multinomial logit
  model}.
\newblock \bibinfo{journal}{Transp. Res. Rec.} \bibinfo{volume}{2672},
  \bibinfo{pages}{35--45}.
\bibitem[{Wang et~al.(2021)Wang, Mo, Hess and Zhao}]{Wang2021}
\bibinfo{author}{Wang, S.}, \bibinfo{author}{Mo, B.}, \bibinfo{author}{Hess,
  S.}, \bibinfo{author}{Zhao, J.}, \bibinfo{year}{2021}.
\newblock \bibinfo{title}{Comparing hundreds of machine learning classifiers
  and discrete choice models in predicting travel behavior: an empirical
  benchmark}.
\newblock \href{http://arxiv.org/abs/2102.01130}{\tt arXiv:2102.01130}.
\bibitem[{Wang et~al.(2020)Wang, Mo and Zhao}]{Wang2019}
\bibinfo{author}{Wang, S.}, \bibinfo{author}{Mo, B.}, \bibinfo{author}{Zhao,
  J.}, \bibinfo{year}{2020}.
\newblock \bibinfo{title}{Predicting travel mode choice with 86 machine
  learning classifiers: An empirical benchmark study}.
\bibitem[{Xu et~al.(2019)Xu, Skoularidou, Cuesta-Infante and
  Veeramachaneni}]{Xu2019}
\bibinfo{author}{Xu, L.}, \bibinfo{author}{Skoularidou, M.},
  \bibinfo{author}{Cuesta-Infante, A.}, \bibinfo{author}{Veeramachaneni, K.},
  \bibinfo{year}{2019}.
\newblock \bibinfo{title}{Modeling tabular data using conditional {GAN}}.
\newblock \bibinfo{journal}{arXiv} .
\bibitem[{Zhang et~al.(2020)Zhang, Ji, Wang and Yang}]{Zhang2020}
\bibinfo{author}{Zhang, Z.}, \bibinfo{author}{Ji, C.}, \bibinfo{author}{Wang,
  Y.}, \bibinfo{author}{Yang, Y.}, \bibinfo{year}{2020}.
\newblock \bibinfo{title}{A customized deep neural network approach to
  investigate travel mode choice with interpretable utility information}.
\newblock \bibinfo{journal}{J. Adv. Transp.} \bibinfo{volume}{2020},
  \bibinfo{pages}{1--11}.
\bibitem[{Zhao et~al.(2020)Zhao, Yan, Yu and Van~Hentenryck}]{Zhao2020}
\bibinfo{author}{Zhao, X.}, \bibinfo{author}{Yan, X.}, \bibinfo{author}{Yu,
  A.}, \bibinfo{author}{Van~Hentenryck, P.}, \bibinfo{year}{2020}.
\newblock \bibinfo{title}{Prediction and behavioral analysis of travel mode
  choice: A comparison of machine learning and logit models}.
\newblock \bibinfo{journal}{Travel Behav. Soc.} \bibinfo{volume}{20},
  \bibinfo{pages}{22--35}.

\end{thebibliography}

\section*{Statements and Declarations}

\begin{itemize}

\item Funding Sources
\\This research did not receive any specific grant from funding agencies in the public, commercial, or not-for-profit sectors.

\item Competing Interests
\\The authors have no relevant financial or non-financial interests to disclose.

\item Author Contribution
\\Conceptualization: Amirhossein Parsi, Melina Jafari, Sina Sabzekar, Zahra Amini; Methodology: Amirhossein Parsi, Melina Jafari, Sina Sabzekar, Zahra Amini; Data data collection and analysis: Amirhossein Parsi, Melina Jafari; Writing-original draft preparation: Melina Jafari, Amirhossein Parsi; Writing-review and editing: Amirhossein Parsi, Zahra Amini; Supervision: Zahra Amini

\end{itemize}

\bio{}

{Amirhossein Parsi}
earned his bachelor's degree in Civil Engineering from Sharif University of Technology in 2021 and is currently pursuing a master's degree in Transportation Engineering at the same institution. His academic focus revolves around Intelligent Transportation Systems (ITS) and sustainable transportation, with an emphasis on integrating machine learning into his research endeavors.

{Melina Jafari} 
completed her bachelor's degree in Civil Engineering at Sharif University of Technology in 2022. She is currently pursuing a master's degree in Transportation Engineering at Sharif University of Technology. Her research interests are Intelligent Transportation Systems (ITS), traffic safety, transportation planning, and traffic engineering.

{Sina Sabzekar} 
obtained his bachelor's degree in Civil Engineering with a minor in Computer Engineering from the University of Tehran in 2020. Currently, he is pursuing a Master of Science degree in Transportation Engineering at Sharif University of Technology. With a keen interest in Intelligent Transportation Systems (ITS) and machine learning applications, his research revolves around addressing transportation problems through innovative approaches.

{Zahra Amini} 
is currently an Assistant Professor at the Department of Civil Engineering, Sharif University of Technology. She completed her bachelor's degree in 2014 and her master's degree in 2015, in Civil Engineering at the University of California, Berkeley. She obtained her Ph.D. in Highway and Traffic Engineering, in 2018 at the University of California, Berkeley. Her research interests are Intelligent Transportation Systems (ITS), traffic theory and control strategies, and transportation system operation and management.

\endbio

\end{document}